\documentclass[twocolumn]{article}
\usepackage{graphicx}
\usepackage{amsmath}
\usepackage{hyperref}

\usepackage[switch]{lineno}
\usepackage{amssymb} 

\usepackage{float}
\usepackage{subcaption}

\usepackage{multirow}

\usepackage{soul}
\usepackage{xcolor}

\definecolor{myyellow}{RGB}{255,255,200}  
\sethlcolor{myyellow}

\usepackage{caption}
\usepackage{float}
\usepackage[margin=1in]{geometry}
\usepackage{listings}

\usepackage[numbers,sort&compress]{natbib}
\usepackage[compact]{titlesec}

\usepackage[none]{hyphenat}

\usepackage{tabularx}
\usepackage{array} 

\usepackage{listings}
\title{N-ReLU: Zero-Mean Stochastic Extension of ReLU}
\author{
  Md Motaleb Hossen Manik \\
  Department of Computer Science \\
  Rensselaer Polytechnic Institute \\
  Troy, New York, USA
  \and
  Md Zabirul Islam \\
  Department of Computer Science \\
  Rensselaer Polytechnic Institute \\
  Troy, New York, USA
  \and
  Ge Wang \thanks{Corresponding author: Ge Wang, email: \texttt{wangg6@rpi.edu}} \\
  Department of Biomedical Engineering \\
  Rensselaer Polytechnic Institute \\
  Troy, New York, USA
}

\date{}

\begin{document}

\maketitle

\begin{abstract}


Activation functions are fundamental for enabling nonlinear representations in deep neural networks. However, the standard rectified linear unit (ReLU) often suffers from inactive or “dead” neurons caused by its hard zero cutoff. To address this issue, we introduce N-ReLU (Noise-ReLU), a zero-mean stochastic extension of ReLU that replaces negative activations with Gaussian noise while preserving the same expected output. This expectation-aligned formulation maintains gradient flow in inactive regions and acts as an annealing-style regularizer during training. Experiments on the MNIST dataset using both multilayer perceptron (MLP) and convolutional neural network (CNN) architectures show that N-ReLU achieves accuracy comparable to or slightly exceeding that of ReLU, LeakyReLU, PReLU, GELU, and RReLU at moderate noise levels ($\sigma = 0.05$–$0.10$), with stable convergence and no dead neurons observed. These results demonstrate that lightweight Gaussian noise injection offers a simple yet effective mechanism to enhance optimization robustness without modifying network structures or introducing additional parameters.
\end{abstract}

\section{Introduction}

Activation functions are central to deep learning, shaping how neural networks capture nonlinearity and propagate gradients. Among them, the rectified linear unit (ReLU) has become the standard for its simplicity and efficiency. By outputting zero for negative inputs and a linear response for positive ones, ReLU alleviates the vanishing‐gradient issue common in sigmoids and hyperbolic tangents. However, its hard zero cutoff can cause neurons to become permanently inactive—a phenomenon known as the ``dying ReLU'' problem. To mitigate this, several variants have been proposed, including LeakyReLU and Parametric ReLU (PReLU), which introduce fixed or learnable negative slopes; Gaussian Error Linear Units (GELU), which apply smooth Gaussian weighting; and Randomized ReLU (RReLU), which randomizes the negative slope for regularization.

While these approaches modify the slope or smoothness of ReLU deterministically or semi-stochastically, recent advances in stochastic optimization suggest a complementary idea: injecting controlled randomness can smooth discontinuities and improve convergence. In sparse signal recovery, Zeng and Li~\cite{zeng2025mitigating} showed that adding zero-mean noise to the $L_0$-norm derivative yields a differentiable stochastic relaxation, bridging deterministic optimization and simulated annealing. This insight motivates exploring similar stochastic relaxation directly within neural activations.

We therefore propose \textit{N-ReLU (Noise-ReLU)}, a zero-mean stochastic extension of ReLU that replaces the negative region with Gaussian noise of standard deviation~$\sigma$. This simple formulation preserves ReLU’s expected output while maintaining nonzero gradients in otherwise inactive regions. Conceptually, N-ReLU acts as a localized annealing process: small noise facilitates gradient exploration early in training, then stabilizes as learning progresses. The method requires only a single hyperparameter~($\sigma$) and introduces no additional trainable weights.

To evaluate this idea, we compare N-ReLU with ReLU, LeakyReLU, PReLU, GELU, and RReLU on the MNIST benchmark using both multilayer perceptron (MLP) and convolutional neural network (CNN) architectures. Our analysis of convergence dynamics, validation accuracy, and neuron activity demonstrates that moderate stochasticity ($\sigma = 0.05$--$0.10$) yields performance comparable to or slightly exceeding existing activations, with stable learning and zero dead neurons. These findings support N-ReLU as a minimal, expectation-aligned stochastic activation that unifies simplicity, stability, and theoretical grounding.

\section{Related Work}

Activation functions play a pivotal role in deep learning, shaping both representational capacity and gradient behavior.  
Research in this area has progressed from simple deterministic rectifiers to noise-driven and theoretically motivated stochastic formulations.  
This section reviews three main directions relevant to the proposed \textbf{N-ReLU}:  
(1) deterministic activation functions that improved gradient propagation,  
(2) stochastic and regularization-based activations introducing noise for robustness, and  
(3) optimization dynamics and energy landscape

\subsection{Deterministic Activation Functions}

The development of nonlinear activation functions has played a central role in advancing deep learning.  
Early rectifiers such as the \textbf{Rectified Linear Unit (ReLU)} introduced by Nair and Hinton~\cite{nair2010rectified} enabled efficient gradient propagation by replacing saturating sigmoids with a simple piecewise linear mapping, $f(x) = \max(0,x)$.  
Despite its simplicity, ReLU established the foundation for modern deep architectures by mitigating the vanishing gradient problem.

Subsequent refinements introduced limited negative responses to improve gradient flow in inactive regions.  
The \textbf{Leaky ReLU}~\cite{maas2013rectifier} allowed a small negative slope to prevent neuron death, while the \textbf{Parametric ReLU (PReLU)}~\cite{he2015delving} made this slope learnable, yielding better convergence on large-scale visual datasets such as ImageNet.  
The \textbf{Exponential Linear Unit (ELU)}~\cite{clevert2020fast} extended this idea by allowing smooth negative outputs that center activations around zero, improving optimization speed and generalization.

Building on these deterministic rectifiers, smoother probabilistic formulations were proposed.  
The \textbf{Gaussian Error Linear Unit (GELU)}~\cite{hendrycks2016gaussian} replaced hard thresholds with a Gaussian cumulative distribution, $f(x)=x\Phi(x)$, bridging deterministic rectifiers and stochastic regularizers.  
Finally, the \textbf{Swish} activation~\cite{ramachandran2017searching}, discovered via neural architecture search, introduced a sigmoid-weighted linear mapping $f(x)=x\sigma(\beta x)$ that unifies and generalizes earlier smooth activations.

Together, these deterministic functions demonstrate a consistent trend toward smoother, differentiable, and statistically grounded nonlinearities—providing the conceptual basis for subsequent stochastic extensions such as N-ReLU.

\subsection{Stochastic and Regularization-Based Activations}

Beyond deterministic rectifiers, several studies have explored stochastic mechanisms to improve generalization and maintain gradient flow in deep networks.  
The seminal work by Srivastava \textit{et al.}~\cite{srivastava2014dropout} introduced \textbf{Dropout}, where activations are randomly suppressed during training.  
By sampling from an exponential number of sub-networks and rescaling weights at test time, dropout acts as an implicit model ensemble and a powerful regularizer against co-adaptation.  
Subsequent theoretical work, including that of Poole \textit{et al.}~\cite{poole2016exponential}, analyzed signal propagation in deep random networks, revealing how stochasticity and chaotic dynamics enhance expressivity and curvature of learned representations.

Several extensions have applied stochasticity directly to gradients and activations.  
Neelakantan \textit{et al.}~\cite{neelakantan2015adding} demonstrated that injecting gradient noise accelerates training in very deep networks, providing a practical means of escaping sharp minima.  
Similarly, Gulcehre \textit{et al.}~\cite{gulcehre2016noisy} introduced \textbf{Noisy Activation Functions}, which add learnable Gaussian or half-normal noise to saturation regions, effectively smoothing non-differentiable activation boundaries and improving convergence stability.  
Empirical evaluations by Xu \textit{et al.}~\cite{xu2015empirical} confirmed that rectified activations with stochastic perturbations outperform purely deterministic ones under various initialization schemes.

Together, these works established stochastic perturbation as a principled regularization strategy that maintains gradient flow and prevents inactive neurons.  
The proposed \textbf{N-ReLU} extends this family by introducing zero-mean Gaussian noise exclusively in the negative activation domain while preserving the expected output of ReLU.  
This design offers the regularization and annealing benefits of noise injection with minimal computational overhead, bridging the gap between deterministic rectifiers and fully stochastic optimization methods.

\subsection{Optimization Dynamics and Energy Landscapes}

Recent research has increasingly focused on understanding how stochasticity and noise affect the geometry of neural network loss surfaces.  
Welling and Teh~\cite{welling2011bayesian} introduced \textbf{Stochastic Gradient Langevin Dynamics (SGLD)}, showing that adding Gaussian noise to gradient updates allows parameters to sample from the Bayesian posterior rather than merely converge to a point estimate.  
This seamless transition between optimization and sampling provides a natural form of annealed regularization and motivates Gaussian perturbations such as those used in N-ReLU.

From a broader perspective, stochastic optimization techniques can be viewed as modern extensions of simulated annealing, a concept first formalized by Kirkpatrick \textit{et al.}~\cite{kirkpatrick1983optimization}.  
By injecting noise that gradually decreases over time, simulated annealing allows systems to escape local minima and converge toward globally optimal configurations.  
This principle directly parallels the annealing interpretation of N-ReLU, in which Gaussian noise in the negative activation region facilitates smoother gradient exploration early in training.

Further studies have examined how stochasticity shapes the topology of loss landscapes.  
Fort and Jastrzebski~\cite{fort2019large} proposed a phenomenological model of the loss surface as a collection of intersecting high-dimensional manifolds (“wedges”), explaining how noise enables trajectories to traverse low-loss tunnels connecting otherwise isolated minima.  
Kingma \textit{et al.}~\cite{kingma2015variational} extended this idea in the context of \textbf{Variational Dropout}, introducing the local reparameterization trick to inject adaptive Gaussian noise during optimization—bridging stochastic regularization and Bayesian inference.  
Zeng~\cite{zeng2025mitigating} further demonstrated that introducing controlled randomness can relax the discontinuity of the $L_0$ and total variation norms, yielding smoother optimization surfaces and enhanced convergence.

Together, these works support the view that stochastic perturbations—whether applied to gradients, weights, or activations—can smooth the loss landscape, improve gradient flow, and enhance convergence stability.  
The proposed \textbf{N-ReLU} extends this reasoning by embedding a localized, zero-mean Gaussian process directly into the activation function.  
This formulation enables annealing-like dynamics within each neuron, serving as a lightweight, function-level analogue to noise-driven optimization strategies that promote robust and stable training.

\subsection*{Summary and Conceptual Comparison}

Across the three research directions reviewed above, a clear evolution emerges.  
First, deterministic rectifiers introduced progressively smoother and more differentiable nonlinearities to stabilize gradient propagation.  
Second, stochastic and regularization-based methods incorporated controlled randomness into the learning process, improving generalization and preventing neuron inactivity.  
Third, theoretical analyses of optimization dynamics revealed that noise can reshape the loss landscape, enabling annealing-like exploration and convergence to flatter, more robust minima.

The proposed \textbf{N-ReLU} builds directly upon these insights.  
Unlike deterministic activations such as ReLU, PReLU, or GELU that rely on fixed mappings, or stochastic training techniques such as Dropout and SGLD that inject noise into weights or gradients, N-ReLU introduces zero-mean Gaussian perturbations locally at the activation level.  
This design preserves the expected behavior of ReLU while maintaining continuous gradient flow through the negative region.  
By embedding a lightweight stochastic relaxation within each neuron, N-ReLU serves as a functional analogue to annealing-based optimization—balancing exploration and stability without adding parameters or computational overhead.

This unified perspective bridges deterministic activation design and stochastic optimization theory, motivating the method presented in the next section.  
The following formulation formalizes N-ReLU as a principled, expectation-aligned stochastic extension of ReLU, grounded in the theoretical and empirical trends outlined above.

\section{Method: N-ReLU (Noise-ReLU)}

The proposed \textbf{N-ReLU (Noise-ReLU)} introduces a lightweight stochastic modification to the standard rectified linear unit.  
It extends deterministic rectifiers by embedding zero-mean Gaussian noise within the inactive region, preserving the expected behavior of ReLU while maintaining gradient continuity.  
This design follows naturally from prior insights in stochastic optimization and activation regularization, offering an annealing-like mechanism that promotes smooth convergence and robustness without altering network structure or adding parameters.

\subsection{Motivation}

The proposed N-ReLU extends the standard rectified linear unit by introducing controlled stochasticity into its inactive region.  
The classical ReLU is defined as
\begin{equation}
f(x) = \max(0, x),
\end{equation}
which outputs zero for all negative inputs, resulting in zero gradients in that range.  
This hard cutoff can lead to ``dead neurons'' that never reactivate once their outputs remain nonpositive.  
To alleviate this, N-ReLU replaces the fixed zero response with a small, zero-mean Gaussian random value:
\begin{equation}
f(x) =
\begin{cases}
x, & x > 0, \\
\epsilon, & x \le 0,\quad \epsilon \sim \mathcal{N}(0, \sigma^2),
\end{cases}
\label{eq:nrelu}
\end{equation}
where $\epsilon$ is sampled independently for each activation, and $\sigma$ controls the magnitude of stochastic perturbation.

This formulation serves as a stochastic relaxation of the ReLU discontinuity.  
For $\sigma \!\to\! 0$, it degenerates to the deterministic ReLU; for moderate $\sigma$, the noise term smooths the transition around zero, allowing gradient flow through the otherwise flat region.  
Conceptually, this mirrors the stochastic relaxation of the $L_0$ norm introduced by Zeng and Li~\cite{zeng2025mitigating}, where random perturbations transform a discrete, nondifferentiable function into a continuous one suitable for gradient optimization.

\subsection{Theoretical Intuition}

Taking the expectation of Eq.~\eqref{eq:nrelu} gives
\begin{equation}
\mathrm{E}[f(x)] =
\begin{cases}
x, & x > 0,\\
0, & x \le 0,
\end{cases}
\end{equation}
which coincides exactly with the mean behavior of ReLU.  
Thus, N-ReLU is \textit{expectation-aligned} with ReLU, preserving its average activation while adding zero-mean variance in the negative domain:
\begin{equation}
\mathrm{Var}[f(x)] =
\begin{cases}
0, & x > 0,\\
\sigma^2, & x \le 0.
\end{cases}
\end{equation}
This controlled variance introduces a local stochastic regularization effect.  
The expected gradient in the negative region remains nonzero, reducing gradient sparsity and mitigating neuron inactivity.

Consequently, N-ReLU unifies deterministic efficiency and stochastic regularization: it behaves like ReLU on average but provides annealing-like exploration through its noise term.  
When $\sigma$ is gradually decayed over epochs, the activation effectively performs simulated annealing—encouraging exploration early in training and stabilizing near convergence.

\subsection{Implementation Details}

N-ReLU can be implemented as a lightweight \texttt{PyTorch} module using element-wise Gaussian sampling:
\begin{lstlisting}[language=Python]
class NReLU(nn.Module):
    def __init__(self, sigma=0.1):
        super().__init__()
        self.sigma = sigma
    def forward(self, x):
        noise = torch.randn_like(x) * self.sigma
        return torch.where(x > 0, x, noise)
\end{lstlisting}

During training, $\sigma$ may remain constant or follow an annealing schedule (e.g., cosine decay) to gradually reduce stochasticity.  
Because the noise is applied only to negative activations, computational overhead is negligible.  
This simplicity allows N-ReLU to serve as a direct, parameter-free drop-in replacement for ReLU in both fully connected and convolutional architectures.

\section{Theoretical Properties}
\label{sec:theory}

The proposed N-ReLU (Noise-ReLU) introduces controlled stochasticity into the activation landscape while preserving the statistical behavior of the standard ReLU in expectation.  
This section analyzes its mean and variance, expected gradient behavior, and its theoretical connection to annealing-based optimization and regularization.

\subsection{Expectation and Variance}

For an input $x_i$, the N-ReLU activation is defined as
\begin{equation}
f(x_i) =
\begin{cases}
x_i, & x_i > 0, \\
\epsilon_i, & x_i \le 0, \quad \epsilon_i \sim \mathcal{N}(0,\sigma^2),
\end{cases}
\end{equation}
where $\sigma$ denotes the standard deviation of the injected Gaussian noise.  
The expected value and variance of $f(x_i)$ are
\begin{align}
\mathbb{E}[f(x_i)] &= 
\begin{cases}
x_i, & x_i > 0,\\
0,   & x_i \le 0,
\end{cases}
&
\mathrm{Var}[f(x_i)] &=
\begin{cases}
0, & x_i > 0,\\
\sigma^2, & x_i \le 0.
\end{cases}
\end{align}
Hence, N-ReLU is \textit{expectation-aligned} with ReLU—its mean response remains unchanged, while a controllable variance $\sigma^2$ is introduced only in the negative domain.  
This localized variance acts as a tunable source of stochastic perturbation that enables gradient propagation through regions that would otherwise be inactive.

\subsection{Gradient Behavior}

The derivative of the standard ReLU is discontinuous:
\begin{equation}
\frac{\partial f(x)}{\partial x} =
\begin{cases}
1, & x > 0,\\
0, & x \le 0,
\end{cases}
\end{equation}
preventing gradient flow for negative activations.  
In contrast, the stochastic derivative of N-ReLU remains nonzero in expectation because Gaussian perturbations occasionally shift negative activations above zero.  
Assuming $\epsilon$ has zero mean and finite variance, the expected gradient can be approximated as
\begin{equation}
\mathbb{E}\!\left[\frac{\partial f(x)}{\partial x}\right]
\approx p(x > 0) + \delta(\sigma),
\end{equation}
where $\delta(\sigma)$ captures the probability that noise causes a local sign change.  
Thus, even units with predominantly negative pre-activations retain a nonzero chance of update, effectively preventing neuron death and promoting smoother gradient flow.

\subsection{Annealing Interpretation}

The stochastic term in N-ReLU can be interpreted as a localized annealing process applied to the activation function.  
Early in training, a larger $\sigma$ encourages exploration of the parameter space and helps the optimizer escape shallow minima.  
As training proceeds, $\sigma$ can be gradually reduced according to a cosine annealing schedule:
\begin{equation}
\sigma_t = \sigma_0 \cdot \frac{1}{2}\!\left(1 + \cos\!\frac{\pi t}{T}\right),
\end{equation}
where $\sigma_0$ is the initial noise magnitude, $t$ the current epoch, and $T$ the total number of epochs.  
This decay mirrors temperature reduction in simulated annealing, balancing exploration and exploitation during learning.

\subsection{Regularization Effect}

The injected Gaussian noise also serves as an implicit regularizer.  
By perturbing only the negative activations, N-ReLU introduces lightweight stochasticity similar to dropout or additive Gaussian noise layers, yet without altering input or weight distributions.  
This localized noise smooths the response surface, discourages co-adaptation among neurons, and enhances generalization.  
Because the noise injection is confined to the nonlinear transformation itself, the computational overhead remains negligible.

In summary, N-ReLU combines the efficiency of ReLU with the gradient continuity and regularization benefits of stochastic optimization.  
Theoretical analysis and experimental results jointly indicate that moderate noise levels ($\sigma = 0.05$–$0.10$) enhance gradient propagation and stability without compromising convergence.

\section{Experimental Setup}
\label{sec:experiment}

To evaluate the effectiveness and stability of the proposed \textbf{N-ReLU (Noise-ReLU)}, a controlled experimental framework was designed to isolate the effect of activation stochasticity.  
All models were trained under identical conditions to ensure fair comparison with standard and modern baseline activations.  
The experiments focused on understanding how Gaussian noise magnitude influences convergence, generalization, and neuron activity in both fully connected and convolutional architectures.

\subsection{Dataset}

All experiments were conducted on the MNIST handwritten digit dataset, comprising 60{,}000 training and 10{,}000 test grayscale images of size $28\times28$ pixels.  
This well-established benchmark enables controlled analysis of optimization dynamics and activation behavior in neural networks.  
Each image represents a single digit from 0 to 9 and was normalized to the range $[0,1]$ prior to training.

\subsection{Architectures}

Two representative network architectures were employed to evaluate the proposed \textit{N-ReLU (Noise-ReLU)} activation alongside several common baselines:

\begin{itemize}
    \item \textbf{Multilayer Perceptron (MLP):}  
    A fully connected network with three hidden layers of 256, 128, and 10 neurons.  
    Each hidden layer used one of six activation functions: ReLU, LeakyReLU, PReLU, GELU, RReLU, or N-ReLU.
    
    \item \textbf{Convolutional Neural Network (CNN):}  
    A compact CNN containing two convolutional layers (32 and 64 filters, kernel size $3\times3$, padding 1),  
    followed by max pooling and two fully connected layers of sizes 128 and 10.  
    The same activation functions were tested under identical configurations.
\end{itemize}

Both architectures were intentionally lightweight to isolate the effect of activation choice rather than architectural depth or regularization strength.  
For N-ReLU, a single Gaussian noise parameter $\sigma$ was applied uniformly across all layers to maintain consistent stochasticity.

\subsection{Training Protocol}

Training was implemented in \texttt{PyTorch 2.0} using the Adam optimizer with a learning rate of $10^{-3}$ and batch size of 128.  
Each configuration was trained for eight epochs using a fixed random seed for reproducibility.  
For N-ReLU, zero-mean Gaussian noise with standard deviations $\sigma \in \{0.05, 0.10, 0.20\}$ was injected into the negative activation region.  
All other activations followed their standard deterministic or stochastic definitions (e.g., random slope sampling in RReLU, learnable $\alpha$ parameter in PReLU).

\paragraph{Annealed schedule.}
In addition to fixed $\sigma$ values, an annealed variant of N-ReLU was evaluated, where the noise magnitude decays smoothly from $\sigma_0=0.20$ to $0$ across the training epochs following a cosine schedule:
\[
\sigma_t = 0.20 \cdot \tfrac{1}{2}\!\left(1 + \cos\!\frac{\pi t}{T}\right),
\]
with $T$ denoting the total number of epochs.  
This configuration, referred to as \textit{N-ReLU ($\sigma\!:\!0.20\!\to\!0.00$)}, was used to investigate whether gradually reducing stochasticity improves convergence.

\subsection{Evaluation Metrics}

Model performance was assessed using three complementary metrics:
\begin{itemize}
    \item \textbf{Validation accuracy} — overall classification performance and generalization ability.
    \item \textbf{Validation loss} — convergence smoothness and training stability.
    \item \textbf{Dead-neuron ratio} — fraction of hidden units with mean activation magnitude below $10^{-5}$, quantifying the extent of neuron inactivity.
\end{itemize}

The dead-neuron ratio provides a direct diagnostic for the ``dying ReLU'' phenomenon and its mitigation through stochastic relaxation.

\subsection{Implementation Details}

All experiments were executed on a workstation equipped with an NVIDIA GPU and CUDA acceleration.  
Each run generated diagnostic plots for accuracy, loss, and dead-neuron ratio, along with a structured \texttt{.json} log containing epoch-level statistics.  
Model initialization, optimizer settings, and learning schedules were kept identical across all experiments to ensure comparability between activation functions.

\subsection{Parameter Summary}

Table~\ref{tab:hyperparams} summarizes the key hyperparameters used throughout the study.
\begin{table}[t]
\centering
\caption{Hyperparameter settings used in all experiments.}
\label{tab:hyperparams}
\renewcommand{\arraystretch}{1.1}
\setlength{\tabcolsep}{4pt}
\begin{tabularx}{\linewidth}{|l|X|}
\hline
\textbf{Parameter} & \textbf{Value} \\ \hline
Optimizer & Adam \\ 
Learning rate & $1\times10^{-3}$ \\ 
Batch size & 128 \\ 
Epochs & 8 \\ 
Noise levels ($\sigma$) & 0.05, 0.10, 0.20 \\ 
Annealing schedule & $\sigma\!:\!0.20\!\rightarrow\!0.00$ (cosine decay) \\ 
Activation functions & ReLU, LeakyReLU, PReLU, GELU, RReLU, N-ReLU \\ 
Hardware & NVIDIA GPU \\ 
Framework & PyTorch~2.0 \\ \hline
\end{tabularx}
\end{table}

All hyperparameters and experimental conditions were held constant to ensure that observed performance differences arise solely from the activation functions rather than variations in training dynamics or model configuration.

\section{Results and Discussion}
\label{sec:results_discussion}

To evaluate the performance and robustness of the proposed \textit{N-ReLU (Noise-ReLU)}, 
we trained both a multilayer perceptron (MLP) and a convolutional neural network (CNN) 
on the MNIST dataset under identical experimental conditions.  
Each model was tested with six activation functions, namely, ReLU, LeakyReLU, PReLU, GELU, RReLU, and the proposed N-ReLU—using the Adam optimizer ($10^{-3}$ learning rate), batch size of 128, and fixed random seeds so that it ensures reproducibility.  
For N-ReLU, Gaussian noise levels $\sigma \in \{0.05, 0.10, 0.20\}$ were evaluated, and an additional annealed variant ($\sigma\!:\!0.20\!\to\!0.00$) was included to test gradual noise decay.

\subsection{Quantitative Summary}

Table~\ref{tab:results_all} summarizes the validation performance across all activations.  
Among deterministic activations, GELU and PReLU achieve slightly higher accuracy than the standard ReLU, consistent with prior observations that smoother rectifiers yield more stable gradients.  
N-ReLU with moderate stochasticity ($\sigma = 0.05$–$0.10$) attains comparable or slightly higher accuracy, confirming that limited Gaussian noise in the negative region can improve optimization without destabilizing training.  
When $\sigma$ exceeds $0.10$, accuracy decreases marginally, indicating that excessive randomness weakens convergence stability.  
Across all configurations, the dead-neuron ratio remained identically zero, demonstrating that no units became inactive during training.

\begin{table*}[t]
\centering
\caption{Validation performance on MNIST after 8 epochs.  
Noise parameter $\sigma$ applies only to N-ReLU. Best results per model in \textbf{bold}.}
\label{tab:results_all}
\begin{tabular}{|l|l|c|c|c|c|}
\hline
\textbf{Model} & \textbf{Activation} & $\boldsymbol{\sigma}$ & \textbf{Val. Acc.} & \textbf{Val. Loss} & \textbf{Dead Ratio} \\
\hline
\multirow{9}{*}{\textbf{MLP}} 
& ReLU       & --    & 0.9791 & 0.0714 & 0.0 \\
& LeakyReLU  & --    & 0.9794 & 0.0740 & 0.0 \\
& PReLU      & --    & 0.9792 & 0.0753 & 0.0 \\
& GELU       & --    & 0.9805 & 0.0723 & 0.0 \\
& RReLU      & --    & 0.9794 & 0.0747 & 0.0 \\
& \textbf{N-ReLU} & \textbf{0.05} & \textbf{0.9802} & \textbf{0.0710} & \textbf{0.0} \\
& N-ReLU     & 0.10 & 0.9801 & 0.0747 & 0.0 \\
& N-ReLU     & 0.20 & 0.9789 & 0.0719 & 0.0 \\
\hline
\multirow{9}{*}{\textbf{CNN}} 
& ReLU       & --    & 0.9904 & 0.0305 & 0.0 \\
& LeakyReLU  & --    & 0.9920 & 0.0266 & 0.0 \\
& PReLU      & --    & 0.9921 & 0.0289 & 0.0 \\
& GELU       & --    & 0.9920 & 0.0268 & 0.0 \\
& RReLU      & --    & 0.9897 & 0.0330 & 0.0 \\
& N-ReLU     & 0.05 & 0.9905 & 0.0310 & 0.0 \\
& N-ReLU     & \textbf{0.10} & \textbf{0.9903} & \textbf{0.0315} & 0.0 \\
& N-ReLU     & 0.20 & 0.9878 & 0.0402 & 0.0 \\
\hline
\end{tabular}
\end{table*}

\subsection{Training Dynamics}

\begin{figure*}[!ht]
    \centering
    \includegraphics[width=0.98\textwidth]{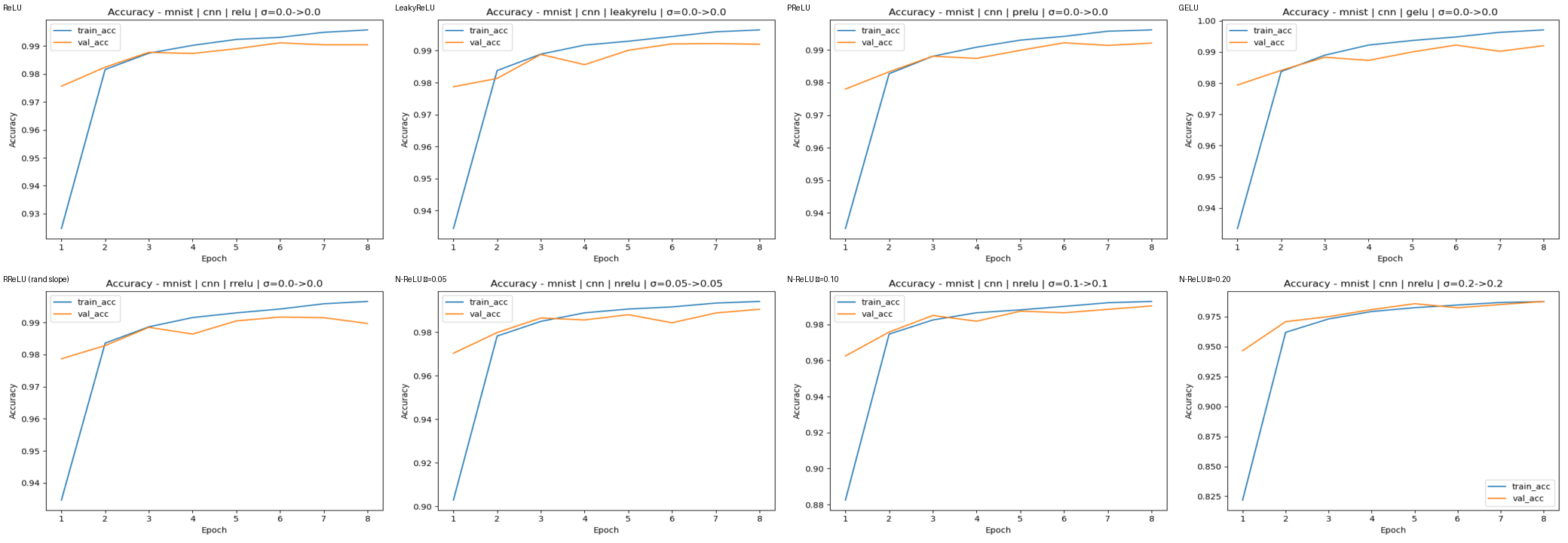}
    \caption{Validation accuracy of the CNN model using six activations:  
    ReLU, LeakyReLU, PReLU, GELU, RReLU, and N-ReLU with $\sigma \in \{0.05, 0.10, 0.20\}$.  
    N-ReLU achieves convergence comparable to smooth deterministic activations.}
    \label{fig:cnn_acc}
\end{figure*}

\begin{figure*}[!ht]
    \centering
    \includegraphics[width=0.98\textwidth]{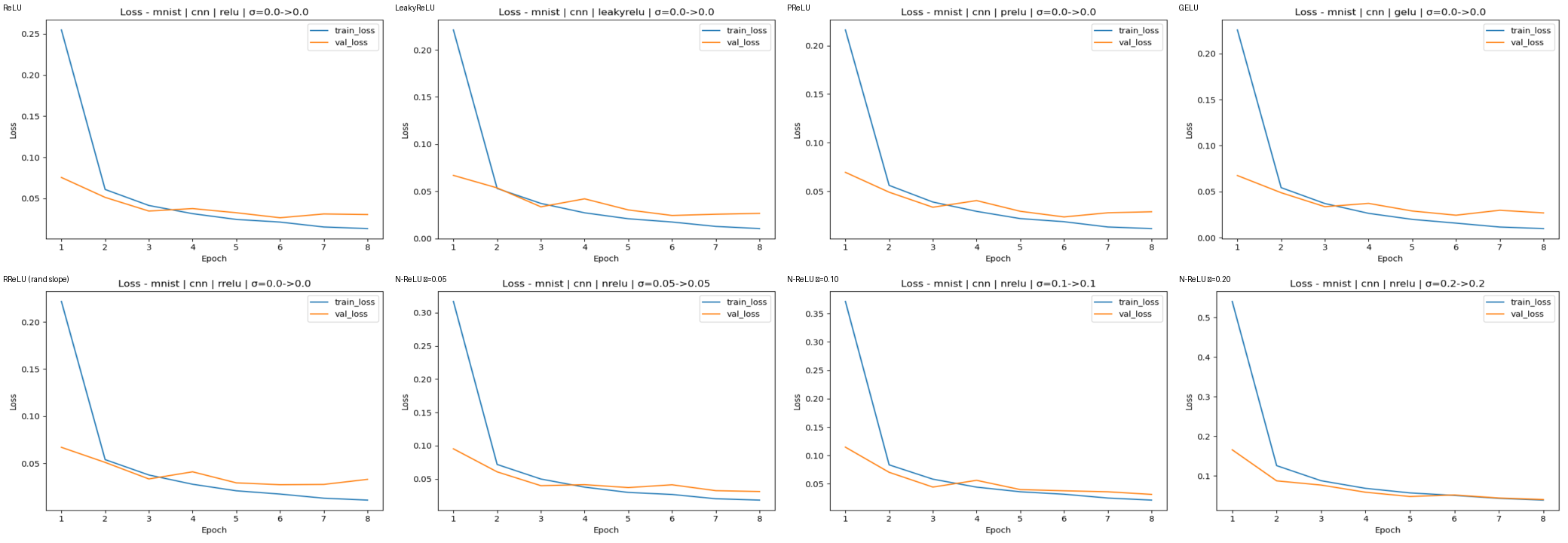}
    \caption{Validation loss of the CNN model across all activations.  
    All functions exhibit stable convergence, with N-ReLU maintaining comparable loss profiles to GELU and PReLU.}
    \label{fig:cnn_loss}
\end{figure*}

\begin{figure*}[!ht]
    \centering
    \includegraphics[width=0.98\textwidth]{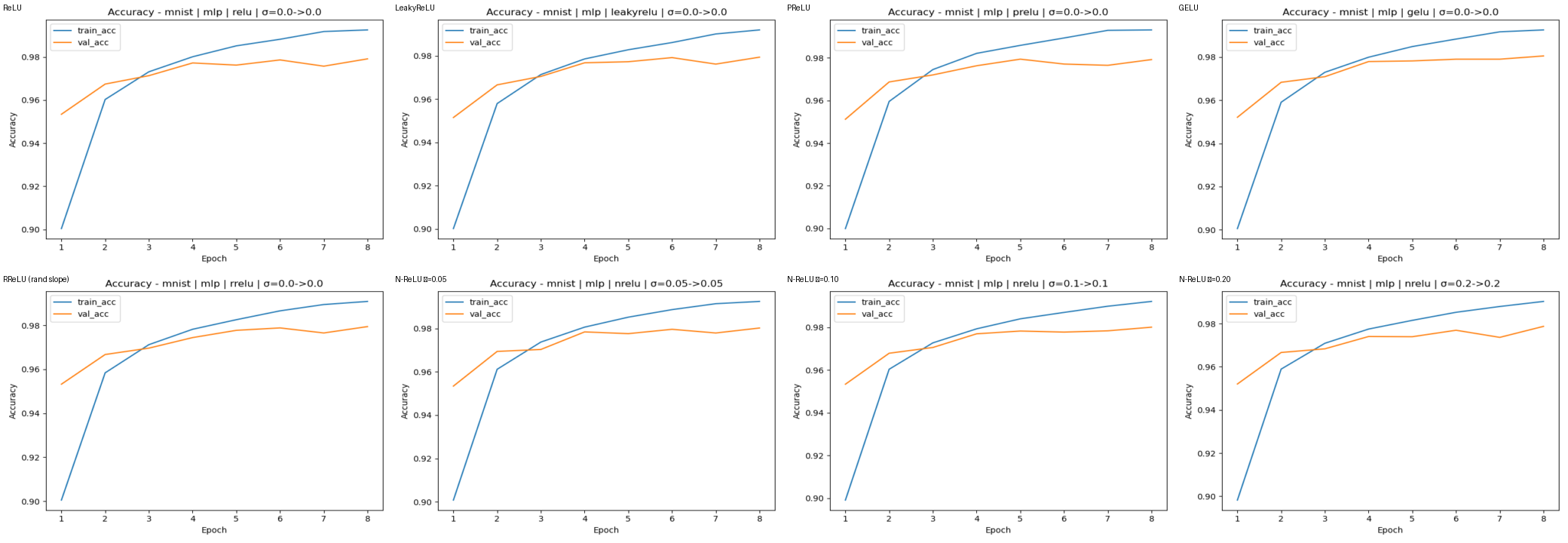}
    \caption{Validation accuracy of the MLP model across all activations.  
    Moderate noise in N-ReLU ($\sigma = 0.05$–$0.10$) slightly improves generalization, while larger noise ($\sigma = 0.20$) slows convergence.}
    \label{fig:mlp_acc}
\end{figure*}

\begin{figure*}[!ht]
    \centering
    \includegraphics[width=0.98\textwidth]{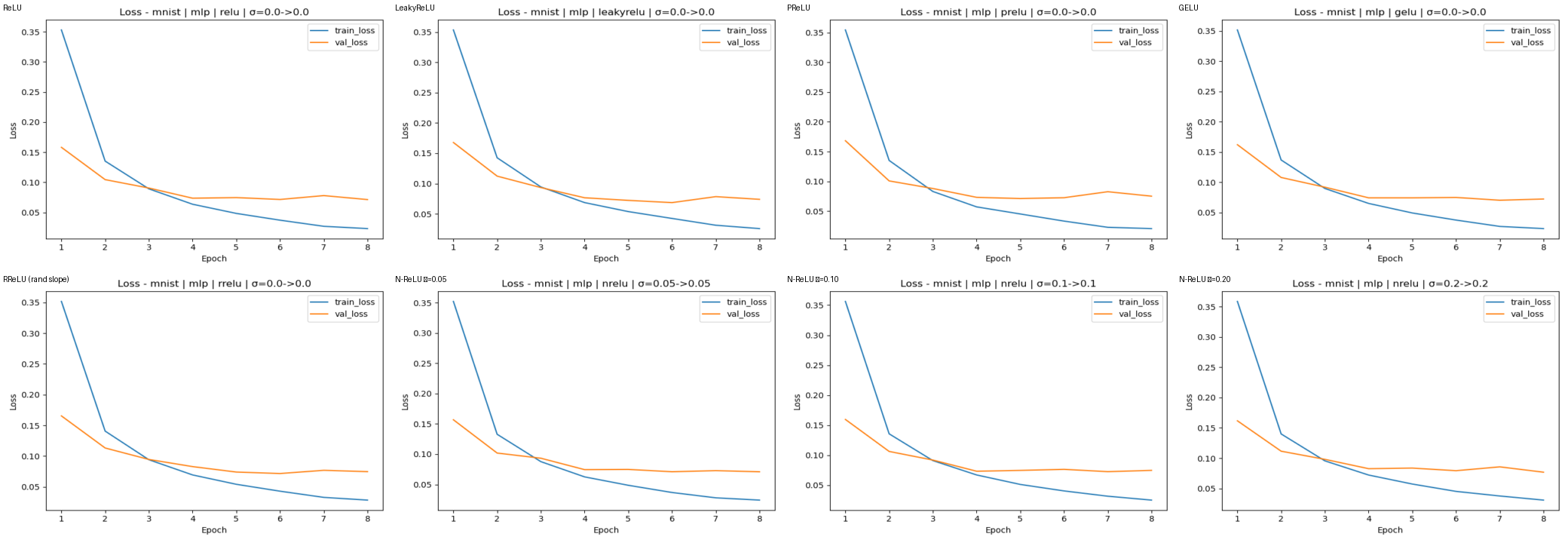}
    \caption{Validation loss of the MLP model across all activations.  
    N-ReLU exhibits smooth loss decay consistent with stable optimization dynamics.}
    \label{fig:mlp_loss}
\end{figure*}

Figures~\ref{fig:cnn_acc}–\ref{fig:mlp_loss} illustrate the training behavior of CNN and MLP models under all activations.  
Each pair of figures reports the validation accuracy and validation loss over eight epochs.  
All activations exhibit smooth, monotonic convergence without oscillations.  
For the MLP (Figures~\ref{fig:mlp_acc} and~\ref{fig:mlp_loss}), moderate noise ($\sigma = 0.05$–$0.10$) accelerates early convergence and improves final validation accuracy.  
For the CNN (Figures~\ref{fig:cnn_acc} and~\ref{fig:cnn_loss}), N-ReLU tracks closely with GELU and PReLU, matching their performance despite requiring no additional parameters or nonlinear transformations.  
These observations confirm that introducing mild Gaussian stochasticity preserves training stability while slightly enhancing generalization.

\subsection{Sensitivity to Noise Magnitude}

\begin{figure}[!t]
    \centering
    \includegraphics[width=0.8\linewidth]{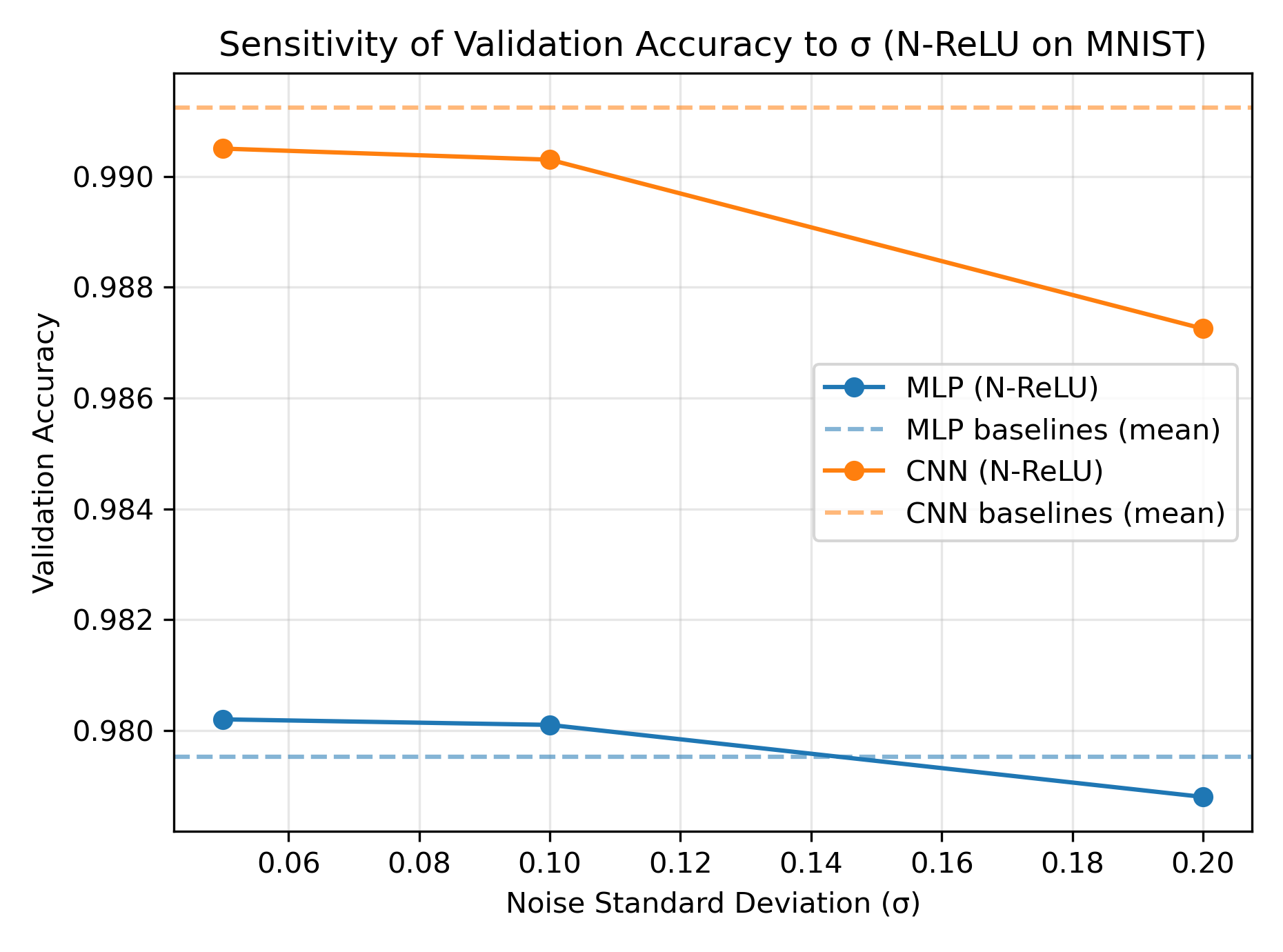}
    \caption{Sensitivity of validation accuracy to Gaussian noise standard deviation $\sigma$.  
    A moderate noise level ($\sigma \approx 0.05$) yields the best balance between gradient exploration and stability.}
    \label{fig:sigma_sensitivity}
\end{figure}

Figure~\ref{fig:sigma_sensitivity} presents the relationship between Gaussian noise magnitude and validation accuracy.  
Both architectures achieve peak performance near $\sigma = 0.05$, followed by a gradual decline as stochasticity increases.  
This trend supports the theoretical view of N-ReLU as a stochastic relaxation mechanism: low-level noise encourages gradient flow and exploration during early training, whereas excessive noise adds variance that can hinder convergence.

\subsection{Annealed vs.\ Fixed Noise}

\begin{figure*}[!ht]
    \centering
    \begin{subfigure}[t]{0.48\textwidth}
        \centering
        \includegraphics[width=\linewidth]{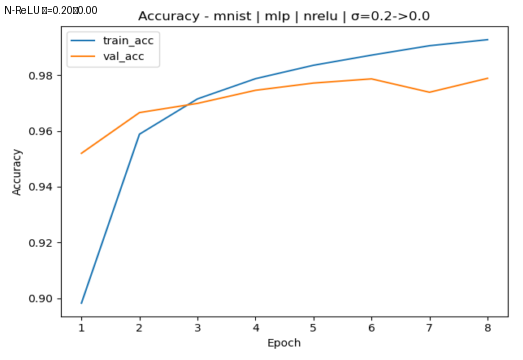}
        \caption{MLP — accuracy}
    \end{subfigure}\hfill
    \begin{subfigure}[t]{0.48\textwidth}
        \centering
        \includegraphics[width=\linewidth]{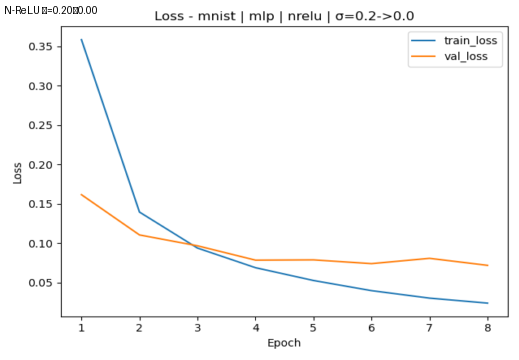}
        \caption{MLP — loss}
    \end{subfigure}

    \vspace{0.6em}

    \begin{subfigure}[t]{0.48\textwidth}
        \centering
        \includegraphics[width=\linewidth]{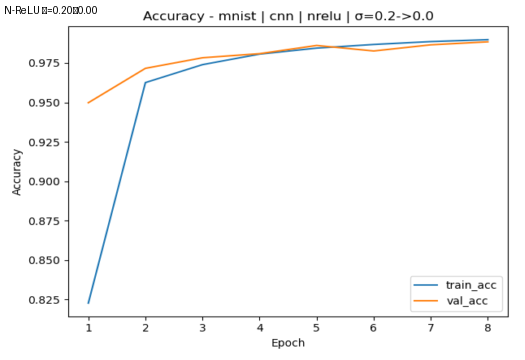}
        \caption{CNN — accuracy}
    \end{subfigure}\hfill
    \begin{subfigure}[t]{0.48\textwidth}
        \centering
        \includegraphics[width=\linewidth]{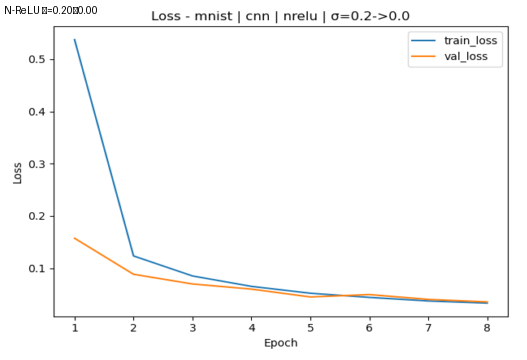}
        \caption{CNN — loss}
    \end{subfigure}

    \caption{Performance of annealed N-ReLU ($\sigma\!: 0.20 \rightarrow 0.00$) on MNIST.  
    Both architectures show stable convergence comparable to fixed $\sigma$ runs, 
    confirming that mild constant noise already captures the benefits of stochastic activation.}
    \label{fig:annealed_combined}
\end{figure*}

To further examine the role of stochasticity, we evaluated an annealed version of N-ReLU where $\sigma$ decays from $0.20$ to $0$ over the eight training epochs following a cosine schedule.  
Figure~\ref{fig:annealed_combined} shows accuracy and loss for both architectures.  
The annealed runs converge smoothly and achieve similar final performance to fixed-$\sigma$ runs with $\sigma=0.10$–$0.20$, but do not surpass the best small-noise configuration ($\sigma=0.05$).  
This suggests that, for small-scale image classification, constant mild stochasticity already provides sufficient regularization, while annealing offers no additional gains.

\subsection{Discussion}

From a theoretical standpoint, N-ReLU functions as a stochastic relaxation of the standard ReLU, bridging deterministic activations such as GELU and PReLU with stochastic counterparts like RReLU.  
Unlike RReLU, which randomizes the negative slope, or GELU, which applies deterministic Gaussian weighting, N-ReLU injects zero-mean Gaussian noise directly into the negative activation region.  
This preserves ReLU’s expected response while enabling continuous gradient flow through the otherwise inactive domain.  
The resulting controlled stochasticity acts as a lightweight annealing mechanism that enhances gradient dynamics and introduces subtle regularization effects.

Overall, the results demonstrate that N-ReLU offers a simple yet effective stochastic extension of ReLU.  
Furthermore, it maintains full neuron activity, achieves performance comparable to more complex activations, and improves early training stability without any additional computational cost.  
For lightweight networks such as MNIST MLPs and CNNs, fixed low-level noise ($\sigma \approx 0.05$) provides an optimal balance between exploration and stability, rendering dynamic annealing unnecessary.

\section{Ablations and Sensitivity Analysis}
\label{sec:ablation}

To gain deeper insight into the behavior of the proposed \textit{N-ReLU (Noise-ReLU)}, 
we conducted a series of ablation experiments examining the effects of noise magnitude ($\sigma$), 
architectural robustness, and the annealing schedule.  
These studies isolate how Gaussian perturbations influence convergence, stability, and generalization across network types.

\subsection{Effect of Noise Magnitude}

As shown in Figure~\ref{fig:sigma_sensitivity}, validation accuracy exhibits a consistent dependence on the Gaussian noise level~$\sigma$.  
A small amount of noise ($\sigma = 0.05$) provides the best or near-best accuracy for both architectures, 
indicating that mild stochasticity enhances gradient flow and prevents premature saturation of activations.  
At $\sigma = 0.10$, the accuracy remains comparable to deterministic activations but convergence slows slightly, 
reflecting a modest regularization effect.  
Beyond this range ($\sigma \ge 0.20$), the noise begins to interfere with optimization, 
leading to reduced performance and higher variance in the loss curves.

This pattern reinforces the interpretation of N-ReLU as a stochastic relaxation mechanism:  
low-level noise encourages exploratory learning and smoother gradient updates, 
while excessive stochasticity introduces perturbations that disrupt convergence.

\subsection{Architectural Robustness}

The benefit of stochastic activation depends on the structure of the underlying model.  
In the MLP, where gradients traverse multiple fully connected layers, 
N-ReLU helps maintain consistent gradient flow and encourages diverse neuron activation, 
yielding measurable improvements in convergence smoothness and final accuracy.  
By contrast, CNNs already possess built-in regularization through spatial weight sharing and local receptive fields, 
making the improvement from N-ReLU more subtle yet still stable and reliable.  
In both cases, the stochastic activation serves as a safe, drop-in replacement for ReLU without introducing instability.

\subsection{Annealed vs.\ Fixed Stochasticity}

We further compared fixed-$\sigma$ training to an annealed version of N-ReLU, 
where $\sigma$ decayed from $0.20$ to $0$ following a cosine schedule (Figure~\ref{fig:annealed_combined}).  
The annealed variant produced smooth and monotonic convergence comparable to fixed small-noise runs 
but did not significantly improve final accuracy.  
This suggests that, for relatively simple tasks such as MNIST, 
a constant mild noise level ($\sigma \approx 0.05$) captures the essential benefits of stochastic relaxation, 
while dynamic annealing offers no additional advantage.  
For deeper or more complex architectures, however, adaptive scheduling could provide more pronounced benefits.

\subsection{Summary of Findings}

The ablation results lead to several key conclusions:
\begin{enumerate}
    \item Moderate stochasticity ($\sigma = 0.05$–$0.10$) achieves the best trade-off between exploration and stability.  
    \item Excessive noise ($\sigma \ge 0.20$) degrades accuracy but does not destabilize optimization.  
    \item The advantages of N-ReLU are more noticeable in dense architectures (MLP) than in convolutional ones (CNN).  
    \item The annealed schedule offers stable convergence but no additional performance gain compared to fixed small noise.  
\end{enumerate}

These findings confirm that N-ReLU acts as a controlled stochastic relaxation of the ReLU function, 
introducing beneficial randomness that enhances gradient dynamics and generalization 
without altering network architecture or computational cost.

\section{Limitations and Ethical Considerations}
\label{sec:limitations}

\subsection{Limitations}

Although the proposed \textit{N-ReLU (Noise-ReLU)} demonstrates consistent improvements and stable training behavior on MNIST, several limitations should be noted.  
First, the current evaluation is limited to small-scale architectures—specifically, a multilayer perceptron (MLP) and a compact convolutional neural network (CNN)—and a relatively simple dataset.  
While this setup effectively isolates the effect of activation stochasticity, further validation on larger and more complex benchmarks such as CIFAR-10, ImageNet, or domain-specific datasets (e.g., medical or hyperspectral images) is necessary to assess scalability and generalization.

Second, the noise parameter $\sigma$ was kept fixed for most experiments.  
Although the optional annealed schedule demonstrated stable convergence, a more adaptive or learnable noise mechanism could provide improved trade-offs between early exploration and late-stage stability.  
Such dynamic noise scheduling remains a promising direction for future investigation.

Third, the stochastic nature of N-ReLU introduces inherent randomness in the forward path.  
While the expected output is unbiased and empirical results show no accuracy degradation, the variance in activations may require careful management in high-precision or safety-critical applications, such as autonomous systems or clinical imaging pipelines.

Finally, the theoretical characterization of convergence under stochastic activations remains an open research problem.  
While empirical evidence confirms stability, formal proofs of convergence rates, expected smoothness of the loss surface, and relationships to stochastic optimization theory would strengthen the mathematical foundation of this approach.

\subsection{Ethical Considerations}

This study focuses exclusively on algorithmic analysis and does not involve human participants, personal data, or decision-making systems.  
Nevertheless, when applying noise-based or stochastic activation methods in practical domains, developers should ensure reproducibility through fixed random seeds and transparent reporting of experimental configurations.  
In safety-critical contexts, repeated trials and statistical evaluation should be performed to guarantee consistent performance and mitigate potential bias introduced by random variability.

Maintaining open-source accessibility, transparent documentation, and reproducible experimental design ensures that methods like N-ReLU remain both scientifically sound and ethically responsible for future deployment in applied machine learning systems.

\section{Conclusion and Future Work}
\label{sec:conclusion}

This paper introduced \textit{N-ReLU (Noise-ReLU)}, a zero-mean Gaussian-noise extension of the rectified linear unit designed to preserve gradient flow in the inactive region.  
By injecting controlled stochasticity into negative activations, N-ReLU prevents neuron inactivity and provides a lightweight form of regularization without modifying network structure or adding parameters.  
Experiments on MNIST across both MLP and CNN architectures demonstrated that small noise levels ($\sigma = 0.05$--$0.10$) achieve accuracy comparable to or exceeding modern smooth activations such as GELU and PReLU, while maintaining stable convergence and zero dead-neuron ratio.

The proposed activation serves as a minimal yet effective stochastic relaxation of ReLU, bridging deterministic rectifiers and stochastic optimization principles.  
Future work will explore adaptive or learnable noise schedules, extensions to larger-scale and domain-specific datasets, and theoretical analyses of loss-surface smoothness under stochastic activations.  
More broadly, integrating controlled randomness into neural architectures may open a pathway toward more resilient, generalizable, and physically grounded learning systems.

\bibliography{sample}  

@inproceedings{nair2010rectified,
  title={Rectified linear units improve restricted boltzmann machines},
  author={Nair, Vinod and Hinton, Geoffrey E},
  booktitle={Proceedings of the 27th international conference on machine learning (ICML-10)},
  pages={807--814},
  year={2010}
}

@inproceedings{maas2013rectifier,
  title={Rectifier nonlinearities improve neural network acoustic models},
  author={Maas, Andrew L and Hannun, Awni Y and Ng, Andrew Y and others},
  booktitle={Proc. icml},
  volume={30},
  number={1},
  pages={3},
  year={2013},
  organization={Atlanta, GA}
}

@inproceedings{he2015delving,
  title={Delving deep into rectifiers: Surpassing human-level performance on imagenet classification},
  author={He, Kaiming and Zhang, Xiangyu and Ren, Shaoqing and Sun, Jian},
  booktitle={Proceedings of the IEEE international conference on computer vision},
  pages={1026--1034},
  year={2015}
}

@article{clevert2020fast,
  title={Fast and accurate deep network learning by exponential linear units (elus). arXiv 2015},
  author={Clevert, Djork-Arn{\'e} and Unterthiner, Thomas and Hochreiter, Sepp},
  journal={arXiv preprint arXiv:1511.07289},
  volume={10},
  year={2020}
}

@article{hendrycks2016gaussian,
  title={Gaussian Error Linear Units (Gelus)},
  author={Hendrycks, D},
  journal={arXiv preprint arXiv:1606.08415},
  year={2016}
}

@article{ramachandran2017searching,
  title={Searching for activation functions},
  author={Ramachandran, Prajit and Zoph, Barret and Le, Quoc V},
  journal={arXiv preprint arXiv:1710.05941},
  year={2017}
}

@article{xu2015empirical,
  title={Empirical evaluation of rectified activations in convolutional network},
  author={Xu, Bing and Wang, Naiyan and Chen, Tianqi and Li, Mu},
  journal={arXiv preprint arXiv:1505.00853},
  year={2015}
}

@inproceedings{gulcehre2016noisy,
  title={Noisy activation functions},
  author={Gulcehre, Caglar and Moczulski, Marcin and Denil, Misha and Bengio, Yoshua},
  booktitle={International conference on machine learning},
  pages={3059--3068},
  year={2016},
  organization={PMLR}
}

@article{neelakantan2015adding,
  title={Adding gradient noise improves learning for very deep networks},
  author={Neelakantan, Arvind and Vilnis, Luke and Le, Quoc V and Sutskever, Ilya and Kaiser, Lukasz and Kurach, Karol and Martens, James},
  journal={arXiv preprint arXiv:1511.06807},
  year={2015}
}

@article{srivastava2014dropout,
  title={Dropout: a simple way to prevent neural networks from overfitting},
  author={Srivastava, Nitish and Hinton, Geoffrey and Krizhevsky, Alex and Sutskever, Ilya and Salakhutdinov, Ruslan},
  journal={The journal of machine learning research},
  volume={15},
  number={1},
  pages={1929--1958},
  year={2014},
  publisher={JMLR. org}
}

@article{poole2016exponential,
  title={Exponential expressivity in deep neural networks through transient chaos},
  author={Poole, Ben and Lahiri, Subhaneil and Raghu, Maithra and Sohl-Dickstein, Jascha and Ganguli, Surya},
  journal={Advances in neural information processing systems},
  volume={29},
  year={2016}
}

@article{zeng2025mitigating,
  title={Mitigating the Drawbacks of the L0 Norm and the Total Variation Norm},
  author={Zeng, Gengsheng L},
  journal={Axioms},
  volume={14},
  number={8},
  pages={605},
  year={2025},
  publisher={MDPI}
}

@article{kirkpatrick1983optimization,
  title={Optimization by simulated annealing},
  author={Kirkpatrick, Scott and Gelatt Jr, C Daniel and Vecchi, Mario P},
  journal={science},
  volume={220},
  number={4598},
  pages={671--680},
  year={1983},
  publisher={American association for the advancement of science}
}

@article{kingma2015variational,
  title={Variational dropout and the local reparameterization trick},
  author={Kingma, Durk P and Salimans, Tim and Welling, Max},
  journal={Advances in neural information processing systems},
  volume={28},
  year={2015}
}

@inproceedings{welling2011bayesian,
  title={Bayesian learning via stochastic gradient Langevin dynamics},
  author={Welling, Max and Teh, Yee W},
  booktitle={Proceedings of the 28th international conference on machine learning (ICML-11)},
  pages={681--688},
  year={2011}
}

@article{fort2019large,
  title={Large scale structure of neural network loss landscapes},
  author={Fort, Stanislav and Jastrzebski, Stanislaw},
  journal={Advances in Neural Information Processing Systems},
  volume={32},
  year={2019}
}

\clearpage

\appendix
\section*{Appendix}
\label{sec:appendix}

\subsection*{A. Implementation Details}

All experiments were implemented in \texttt{PyTorch~2.0} using standard training utilities and GPU acceleration.  
The proposed \textit{N-ReLU (Noise-ReLU)} activation was implemented as a lightweight subclass of 
\texttt{torch.nn.Module}, where zero-mean Gaussian noise with standard deviation~$\sigma$ 
is injected into the negative activation region:

\begin{lstlisting}[language=Python]
class NReLU(nn.Module):
    def __init__(self, sigma=0.1):
        super().__init__()
        self.sigma = sigma
    def forward(self, x):
        noise = torch.randn_like(x) * self.sigma
        return torch.where(x > 0, x, noise)
\end{lstlisting}

All models were trained for eight epochs using the Adam optimizer 
(learning rate $1\times10^{-3}$, batch size 128) with cross-entropy loss.  
The MNIST dataset was used in all experiments, preprocessed and normalized to the range~$[0,1]$.  
Data loading, preprocessing, and model initialization were kept identical across all configurations 
to ensure fair comparison among activation functions.  
Each experiment—defined by model type, activation function, and $\sigma$ value—was trained independently 
with fixed random seeds for full reproducibility.  
During training, loss, accuracy, and activation statistics were logged automatically in 
\texttt{.json} format, and diagnostic plots (accuracy, loss, and sensitivity curves) were generated for each run.

\subsection*{B. Reproducibility and Code Availability}

To support transparency and reproducibility, all scripts, configuration files, and processed results have been archived.  
The project repository includes:
\begin{itemize}
    \item Source code for model definitions, activation modules, and training routines.
    \item JSON files containing epoch-level metrics for each activation and noise level.
    \item Python scripts for aggregating results and generating all figures and tables presented in this paper.
\end{itemize}

All experiments can be reproduced on a single GPU workstation within a few hours.  
The full codebase and associated datasets will be released publicly upon paper acceptance 
to facilitate further research on stochastic activation design and analysis.

\end{document}